\documentclass[referencestyle]{techreport}
\usepackage[utf8]{inputenc}
\usepackage{amsmath,amssymb,amsfonts,bm}
\usepackage[numbers,sort&compress]{natbib}
\usepackage{booktabs,tabularx,array,multirow,makecell}
\usepackage{colortbl}
\usepackage{enumitem}
\usepackage{xspace}
\usepackage{url}
\usepackage{float}
\usepackage{adjustbox}
\usepackage{multicol}
\usepackage{tikz}
\usepackage{wrapfig}
\usetikzlibrary{arrows.meta,positioning,fit,calc,backgrounds,shapes.geometric,decorations.pathreplacing}

\hypersetup{
  pdftitle={Visko Orbis 1.0},
  pdfauthor={Team Visko},
  pdfsubject={Real-Time Interactive Long Video Generation}
}

\newcommand{\model}{\textsc{Visko Orbis 1.0}\xspace}

\newcolumntype{Y}{>{\raggedright\arraybackslash}X}
\newcolumntype{C}[1]{>{\centering\arraybackslash}p{#1}}
\newcolumntype{L}[1]{>{\raggedright\arraybackslash}p{#1}}

\setlist[itemize]{leftmargin=1.35em,itemsep=0.22em,topsep=0.3em}
\setlist[enumerate]{leftmargin=1.55em,itemsep=0.22em,topsep=0.3em}

\definecolor{DataTeal}{HTML}{0E9384}
\definecolor{ModelPurple}{HTML}{7F56D9}
\definecolor{SystemOrange}{HTML}{EF6820}
\definecolor{EvalGreen}{HTML}{039855}
\definecolor{RankFirstGreen}{HTML}{166534}
\definecolor{RankSecondGreen}{HTML}{86CFA5}
\definecolor{RankThirdGreen}{HTML}{DDF3E5}

\newcommand{\rankfirst}[1]{{\cellcolor{RankFirstGreen}\color{white}#1}}
\newcommand{\ranksecond}[1]{{\cellcolor{RankSecondGreen}\color{black}#1}}
\newcommand{\rankthird}[1]{{\cellcolor{RankThirdGreen}\color{black}#1}}

\tikzset{
  trbox/.style={draw=ViskoLine,rounded corners=2mm,thick,fill=white,align=center,inner sep=4.5pt,font=\sffamily\small},
  trsmall/.style={draw=ViskoLine,rounded corners=1.5mm,fill=white,align=center,inner sep=3.5pt,font=\sffamily\scriptsize},
  trarrow/.style={-{Latex[length=2.4mm]},thick,draw=ViskoMuted},
  trdashed/.style={-{Latex[length=2.2mm]},thick,dashed,draw=ViskoMuted}
}

\title{Visko Orbis 1.0: A Live Model for Real-Time\\Interactive Long Video Generation}
\author{Team Visko}
\date{}

\abstract{
We present \model, a Live Model for real-time, interactive long video generation. Users can change the prompt at any moment during generation, and the update becomes visible in real time. \model supports long-form text-to-video, image-to-video, and video continuation, with multilingual prompts and prompt switching while generation is in progress. A bounded multi-scale memory preserves subjects, scenes, and style across chunks, sustaining hour-scale rollouts without evident quality or color drift. The generator is factorized causally in time, matching the causal structure of physical dynamics, and is aligned with a latent world-model reward for predictive consistency. Built on a distilled chunk-wise streaming generator and a streaming video upscaler, \model delivers 4K video generation at 24 FPS in real time, using an optimized GPU serving engine. In quantitative evaluations, \model achieves the best DOVER aesthetic and technical scores and the best VideoAlign visual and motion quality, and leads three physical-plausibility protocols (VideoPhy-2, Physics-IQ, and VBench-2.0 Physics); in long-form Arena comparisons, it obtains the highest overall-preference and temporal-stability ratings among all the state-of-the-art real-time interactive video generation systems.
}

\begin{document}
\maketitle
\pagestyle{reportbody}

\begin{figure}[H]
\centering
\includegraphics[width=\linewidth]{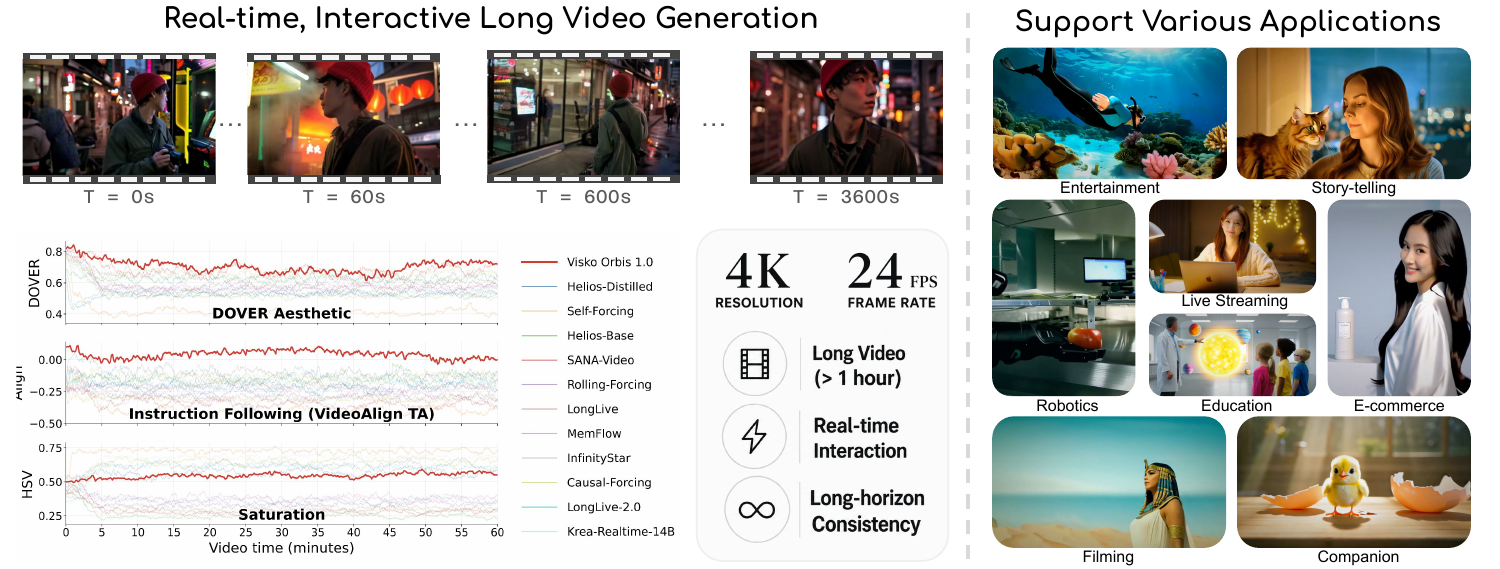}
\caption{Overview of \model as a real-time, interactive long-video generation system. \textbf{Left:} Representative frames at 0, 60, 600, and 3600 seconds illustrate an hour-long rollout, while full-stream DOVER aesthetic, VideoAlign instruction-following, and HSV saturation trajectories track visual quality, prompt alignment, and color stability over time; red denotes \model and lighter curves denote comparison systems. The system delivers 4K output at 24 FPS with real-time interaction and long-horizon consistency. \textbf{Right:} Illustrative applications span entertainment, storytelling, live streaming, robotics, education, e-commerce, filming, and virtual companionship.}
\label{fig:live-model-teaser}
\end{figure}

\section{Introduction}

Video generation has advanced rapidly in visual fidelity, motion realism, instruction following, and multimodal conditioning. Systems such as Sora~\cite{sora2024}, Veo~\cite{veo2024}, Seedance~\cite{seedance2025}, Kling~\cite{klingomni2025}, Wan~\cite{wan2025}, HunyuanVideo~\cite{hunyuanvideo2024}, and LongCat-Video~\cite{longcatvideo2025} now produce short videos of striking quality. Most of them, however, generate offline: a user submits a prompt, waits, and receives a finished clip that can no longer be changed. This paradigm serves bounded content creation well, but it excludes the experiences in which video is consumed while it is being made---live streams, games, virtual characters, interactive stories---where content must keep playing and keep responding to its audience. Such experiences demand a model that generates in real time, sustains long horizons, and stays under continuous user control.

Existing work addresses parts of this demand, but no system covers it end to end. (1)~\emph{Causal streaming.} CausVid~\cite{causvid2025} makes bidirectional video diffusion causal and streamable through few-step distillation and KV caching, and Self Forcing~\cite{selfforcing2025}, Rolling Forcing~\cite{rollingforcing2025}, and Causal Forcing~\cite{causalforcing2026} further study exposure mismatch in causal distillation; these models stream, but live user control is not their focus. (2)~\emph{Interaction.} Oasis~\cite{oasis2024} couples autoregressive generation with a low-latency interactive stack, LongLive~\cite{longlive2025} studies sequential prompt updates over minute-scale rollouts, Matrix-Game 2.0~\cite{matrixgame2025} streams action-conditioned worlds, Krea Realtime~\cite{krea2025} and Helios~\cite{helios2026} pursue real-time long-form generation, and Vidu S1~\cite{vidus12026} supports speech-guided character interaction. (3)~\emph{Long-horizon stability.} FramePack~\cite{framepack2025} and FAR~\cite{far2025} compress long contexts under fixed token budgets, and MemFlow~\cite{memflow2025}, VideoSSM~\cite{videossm2025}, FadeMem~\cite{fademem2026}, and Echo-Infinity~\cite{echoinfinity2026} explore retrieval, hybrid state-space memory, distance-aware consolidation, and learned evolving memory; even so, small errors recirculate through generated history and can accumulate into drift. (4)~\emph{Serving.} Real-time delivery is as much an inference problem as a modeling problem: without state reuse, parallel execution, and streaming decode and upscaling, a capable model may still be too slow to watch. An interactive live model must hold all four together: accept updates while a rollout is active, preserve useful visual state across chunks, deliver frames progressively, and measure its response at the visible output rather than at an internal model boundary.

These requirements reflect a broader shift that we have articulated as \textbf{Live Models}~\cite{visko2026livemodels}: foundation models that execute as persistent processes rather than answering bounded queries. A Live Model generates in real time, under the clock of a world in motion; interacts continuously, folding new input into a process already underway; and maintains an ongoing existence, carrying state forward as long as it runs. For video generation, this contract fixes the system boundary at delivered video: a rollout is a stateful process in which generation history is carried between chunks, prompt updates are timestamped against the output clock, and new conditions apply without reinitializing the rollout. Evaluation adopts the same boundary, reporting quality, continuity across updates, and long-horizon drift over complete delivered streams.

We introduce \textbf{\model}, our first realization of this paradigm: an interactive Live Model supporting long-form text-to-video (T2V), image-to-video (I2V), video-to-video (V2V) continuation, multilingual prompting, and in-generation prompt switching. At its core, a latent-video model generates successive temporal chunks under a single conditional formulation. The chunk-wise causal factorization is chosen not only for latency but because it mirrors the causal, effectively Markovian structure of the physics being generated (Section \ref{ssec:formulation}). Each chunk reads the active event condition, optional visual references, and a bounded multi-scale memory of the preceding rollout; the resulting state is carried into the next chunk. This design separates the persistent visual state of a session from the user's transient intent, so every entry mode shares the same future-generation process.

The result is a system that a user can steer while it runs. A prompt update becomes visible in the output in under one second on average, and decoded chunks are delivered progressively rather than after the full video completes. Native $832\times480$ generation, reference-aware streaming super-resolution, and a multi-GPU serving pipeline together deliver 4K video at 24 frames per second.

We summarize the key technical contributions of \model as follows:

\begin{itemize}
  \item \textbf{Unified live-video formulation.} We formulate generation as a persistent, chunk-wise latent-flow process in which visual state is carried across chunks and time-indexed instructions can change without restarting the rollout. The same causal continuation interface supports T2V, I2V, and V2V generation.

  \item \textbf{Event-aligned data and progressive streaming training.} A multi-stage data engine combines safety and quality filtering, distribution balancing, and both single-event and temporally localized multi-event captions. We first learn a bidirectional short-video prior and then adapt it to history-conditioned streaming through single-event, event-aligned long-video, and human-curated training stages.

  \item \textbf{Bounded memory for long-horizon generation.} Recent history is retained at high latent resolution, older spans are progressively compressed, and information leaving the scheduled window is consolidated into a fixed-capacity learned state. Structured history perturbations and rollout-calibrated augmentation expose the model to recurrent generation errors while keeping memory and per-chunk computation bounded.

  \item \textbf{Causal factorization as a physics prior.} We argue that chunk-wise
   causal generation poses the physics simulation problem as an initial-value problem with time-stepping solution, matching
   the Markovian structure of physics, whereas joint bidirectional
   denoising resembles space–time collocation method without the neccessary iterative solver
   that makes collocation accurate. 

  \item \textbf{Few-step flow post-training with physics-aware alignment.} Guidance distillation removes the unconditional serving branch; history-conditioned trajectory consistency provides a stable few-step initialization; and self-forcing distribution matching adapts the student to its own autoregressive histories. Group-relative reinforcement learning subsequently optimizes calibrated visual-quality, motion-quality, and text--video-alignment rewards. In addition, a latent world model scores candidate futures for predictive consistency, steering generation toward physically plausible motion and object dynamics.

  \item \textbf{Live control and inference alignment.} A rolling prompt summary, asynchronous prompt encoding, chunk-boundary condition updates, and versioned state invalidation let new instructions affect future video while preserving established session context. Conservative content-adaptive drift controls further stabilize long rollouts at inference time.

  \item \textbf{Streaming high-resolution system co-design.} Versioned state reuse, compiled and fused execution, batch-one sequence parallelism, and overlap across generation, decoding, restoration, encoding, and delivery form a single progressive pipeline. A reference-aware, single-refinement video super-resolution model uses temporally distilled decoding, conservative anchor refresh, and tiled distributed execution to transform native $832\times480$ chunks into progressively delivered 4K video.
\end{itemize}

Evaluation treats complete events, rather than selected short clips or individual frames, as the unit of analysis. On the reported one- to three-minute outputs, \model obtains the highest DOVER aesthetic and technical scores and the highest VideoAlign visual- and motion-quality scores among the compared outputs; prompt-aware raw scores are interpreted descriptively because prompt suites are unmatched. In the long-form Arena study, \model achieves the highest overall-preference and temporal-stability Elo point estimates, while the one-hour diagnostic in \cref{fig:live-model-teaser} tracks aesthetic quality, instruction following, and saturation throughout a substantially longer rollout. Three physical-plausibility protocols—long-video VideoPhy-2 and
 Cosmos-Reason scoring, Physics-IQ, and VBench-2.0 Physics—test the
 causal-factorization argument of Section \ref{ssec:formulation} at the delivered-video
 boundary. Full results and qualifications appear in \cref{sec:evaluation}.

The method is organized into three sections. Data describes the governed training data and temporal captions; Model follows the progressive path from bidirectional pretraining through streaming adaptation, post-training, and super-resolution; and Inference describes online prompt control and the delivered-video runtime.

\section{Data}
\label{sec:data}

\begin{figure}[!t]
\centering
\includegraphics[width=0.92\linewidth]{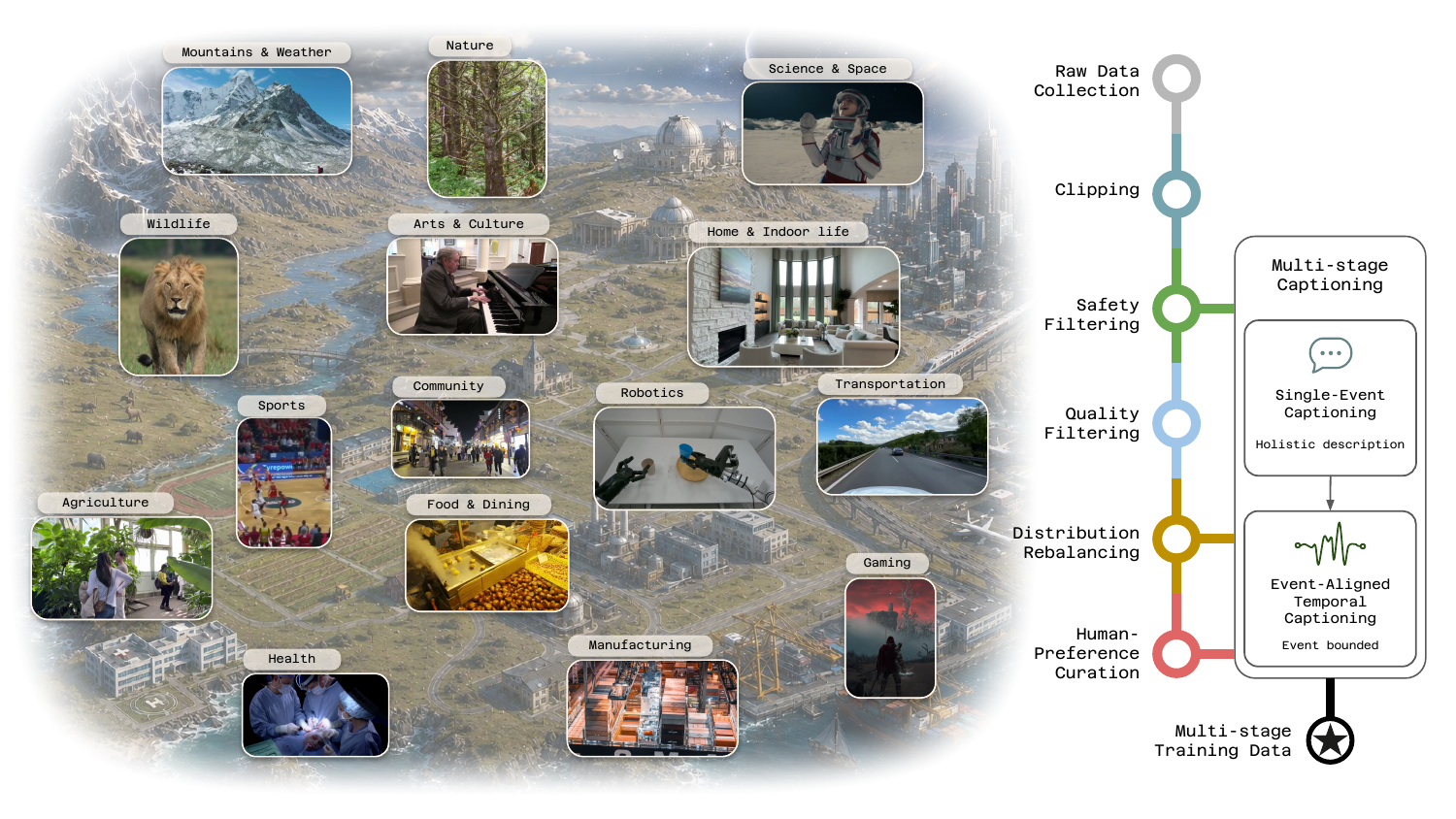}
\caption{\textbf{Multi-stage data curation and captioning pipeline.} The raw video collection spans diverse natural, human-centered, indoor, industrial, transportation, health, sports, and digital domains. Raw videos undergo shot-aware clipping, safety and quality filtering, distribution rebalancing, and human-preference curation. In parallel, every safety-passed clip receives a single-event caption and, where applicable, event-aligned temporal captions, producing governed data subsets for multi-stage training.}
\label{fig:data-engine}
\end{figure}

Data quality is a primary scaling dimension for \model. We built the multi-stage data curation and captioning pipeline illustrated in \cref{fig:data-engine} to filter, annotate, rebalance, and preprocess a large-scale raw video collection. The system produces progressively narrower data tiers for broad-coverage pretraining, distribution-balanced mid-training, and high-quality fine-tuning rather than treating every surviving clip as interchangeable. Large-scale video datasets and generation systems similarly identify filtering, recaptioning, distribution control, and high-quality fine-tuning as central parts of the training recipe~\cite{internvid2023,panda70m2024,stablevideodiffusion2023,cogvideox2024,moviegen2024}.

\paragraph{Raw Video Collection.}

The raw video collection covers a wide range of real-world and generated visual domains. At a high level, these include the natural world and science; wildlife and animals; people, community, education, culture, sports, food, and health; urban, indoor, agricultural, industrial, and transportation environments; and creative, craft, digital, and gaming content. We consolidate exact and near-duplicate material before further processing and retain source and technical metadata for governance and analysis~\cite{datasheets2018,datacards2022}. Detailed source proportions and raw-data totals are maintained in the internal data manifest.

\paragraph{Shot-Aware Video Clipping.}

Raw videos vary from short recordings to long, heterogeneous programs and are not directly suitable as uniform training examples. We first detect shot boundaries from visual discontinuities and temporal change, following established shot-transition detection practice~\cite{transnetv22020}. The resulting shots are then assembled or subdivided according to semantic continuity, content richness, action completeness, and camera behavior, producing clips between 3 and 240 seconds. Clipping is not a fixed-window operation. Adjacent shots may remain together when they form a coherent event or narrative unit, while a long shot can be divided around a meaningful action or scene change.

\paragraph{Safety Filtering.}

Every candidate clip passes a layered safety and privacy filter before entering a training pool. Automated classifiers and rule-based checks screen for sexual or NSFW content, sexual exploitation or endangerment involving minors, graphic violence and gore, self-harm, hateful or extremist material, dangerous or illegal activity, sensitive personal information, and identity-sensitive content. Category-specific checks distinguish high-risk depictions from legitimate documentary, educational, scientific, or medical context where appropriate. High-confidence policy violations are excluded. Ambiguous or high-impact cases are quarantined for additional review rather than accepted through a single permissive threshold.

\paragraph{Quality Filtering.}

Long-horizon generation amplifies defects that may be tolerable in an isolated short clip: color-channel errors can become persistent casts, weak motion can produce static rollouts, and temporal jitter can propagate across later chunks. Safety-passed clips therefore undergo a cascaded assessment of structural integrity, spatial and technical quality, temporal quality, composition, and semantic usefulness, extending the multi-stage filtering practice used in large-scale video diffusion pipelines~\cite{stablevideodiffusion2023,cogvideox2024,moviegen2024}. Representative signals include decode and frame-rate integrity, corrupted or frozen frames, sharpness, exposure, color stability, compression, aesthetic quality, motion amplitude, flicker, jitter, camera stability, subject visibility, framing, event clarity, content richness, and captionability. Thresholds are calibrated by content and source type rather than applied universally, since animation, cinematic footage, scientific visualization, screen content, handheld video, sports, and low-light scenes have different expected distributions.

\paragraph{Distribution Rebalancing.}

Aggressive quality filtering alone can yield a clean but narrow training set: common source types, static scenes, or visually polished categories may dominate even when rarer events are important for model behavior. We maintain a coverage ledger over semantic domain, event and transition type, duration, motion regime, camera behavior, visual style, language context, and whether subjects or environments should persist across a boundary. Rebalancing operates on this ledger after quality scoring. Overrepresented strata are downweighted, underrepresented but valid strata receive higher sampling priority, and near-duplicate semantic clusters are prevented from consuming disproportionate training capacity. Balancing is stage specific: broad pretraining retains long-tail coverage, mid-training applies stronger distribution control, and fine-tuning prioritizes quality without collapsing event diversity.

\paragraph{Human-Preference Curation.}

The highest-quality tier is selected through human annotation and verification. Reviewers assess visual and technical quality, composition, semantic clarity, motion naturalness, temporal coherence, and the presence of subtle defects that automated metrics may miss. Human review also checks borderline automatic decisions and provides calibrated preference judgments among otherwise strong candidates. Only verified examples enter the human-preference pool used for high-quality fine-tuning. Reviewer feedback is aggregated into a versioned decision record and is also used to audit automatic quality signals, identify systematic false positives or false negatives, and propose future filter recalibration. This stage is intentionally narrower than the preceding quality-filtered pool: its purpose is to establish a concentrated quality ceiling, not to reproduce the full raw-data distribution.

\paragraph{Multi-Stage Captioning.}

All clips that pass the safety filter are captioned so that the resulting governed data pool can support different stages of training. Caption generation is asynchronous and versioned independently from quality scores and sampling manifests. Two complementary annotation modes are produced:

\begin{enumerate}
  \item \textbf{Single-event captioning.} Clips of up to 10 seconds receive a holistic caption describing the visible subjects, actions, environment, camera behavior, style, and temporal evolution as one coherent event, building on scalable multi-teacher and model-assisted video recaptioning practice~\cite{internvid2023,panda70m2024,cogvideox2024}.
  \item \textbf{Event-Aligned Temporal Captioning.} Every clip between 3 and 240 seconds undergoes event detection. Captions are generated from the visual evidence inside each detected event interval rather than from an undifferentiated full-video summary. The resulting record contains ordered event boundaries, local event captions, and persistent subject, scene, and style attributes that may span multiple intervals, following the temporally localized supervision used in dense event captioning and structured long-video datasets~\cite{activitynetcaptions2017,vid2seq2023,miradata2024}.
\end{enumerate}

Short clips can carry both views: a single-event caption for compact supervision and event-boundary metadata for a common temporal schema. For longer clips, interval-local captions prevent an action described late in a paragraph from being assigned to an earlier generation window.

These annotations support progressively curated subsets of the same governed training data. Broad-Coverage Pretraining uses the safety-passed data to establish semantic and visual coverage; Distribution-Balanced Mid-training uses a quality-filtered and rebalanced mixture to strengthen underrepresented domains, motion regimes, and event transitions; and High-Quality Fine-tuning concentrates human-verified examples with verified single- and multi-event captions.

\section{Model}
\label{sec:model}

\begin{figure}[H]
\centering
\resizebox{\linewidth}{!}{%
\begin{tikzpicture}[node distance=6mm and 5mm]
  \node[trbox,minimum width=25mm,minimum height=15mm,fill=teal!5,draw=DataTeal] (bidir) {Bidirectional pretraining\\\scriptsize 1--5 s videos};
  \node[trbox,right=of bidir,minimum width=27mm,minimum height=15mm,fill=blue!4,draw=ViskoBlue] (stream) {Streaming adaptation\\\scriptsize 3--10 s videos};
  \node[trbox,right=of stream,minimum width=27mm,minimum height=15mm,fill=purple!4,draw=ModelPurple] (mid) {Event mid-training\\\scriptsize single + multi-event};
  \node[trbox,right=of mid,minimum width=27mm,minimum height=15mm,fill=orange!5,draw=SystemOrange] (fine) {Quality fine-tuning\\\scriptsize human-curated quality tier};
  \node[trbox,right=of fine,minimum width=28mm,minimum height=15mm,fill=green!4,draw=EvalGreen] (post) {Post-training\\\scriptsize distillation + GRPO};

  \draw[trarrow] (bidir) -- (stream);
  \draw[trarrow] (stream) -- (mid);
  \draw[trarrow] (mid) -- (fine);
  \draw[trarrow] (fine) -- (post);



\end{tikzpicture}%
}
\caption{\textbf{Progressive model training.} A short-video bidirectional model is first adapted to chunk-wise streaming generation, then specialized with single- and multi-event data, a human-curated quality tier, few-step distillation, and group-relative reward alignment.}
\label{fig:training-roadmap}
\end{figure}
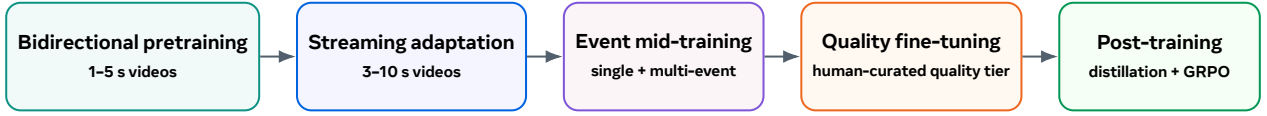

\model is a conditional latent-video model trained progressively from bounded
bidirectional generation to interactive, chunk-wise autoregressive streaming.
This section first states the generation contract and why it is factorized
causally, then follows the order in which capabilities are acquired: the
progressive training path from short-video pretraining to human-curated
fine-tuning, few-step distillation and reward alignment, and a dedicated
super-resolution model. The complete progression is summarized in
\cref{fig:training-roadmap}.

\subsection{Formulation}
\label{ssec:formulation}

\paragraph{Model Formulation.}

Let $z_k$ denote the $k$-th latent-video chunk, $H_k$ the bounded visual history available before that chunk, $c_k$ the text instruction active at the chunk boundary, and $r_k$ optional visual context carried from an input image or video prefix. Instructions are treated as external controls rather than observational random variables: after earlier output has been committed, the user or control policy supplies $c_k$, and the model draws the next chunk from $p_\theta(z_k\mid H_k,c_k,r_k)$. Composing these kernels in generation order defines the causal rollout law even when later instructions are selected adaptively; neither the current kernel nor the state update can access a future instruction or chunk.

The rollout begins from $(H_1,r_1)$. T2V uses empty visual inputs, I2V initializes $r_1$ from the supplied image while leaving $H_1$ empty, and V2V initializes both from the encoded prefix. After $z_k$ is committed, a deterministic causal update constructs the state for $z_{k+1}$. Consequently, later prompt updates affect only uncommitted chunks and never revise delivered output.

\begin{wrapfigure}{r}{0.46\textwidth}
\centering
\vspace{-\intextsep}
\includegraphics[width=\linewidth]{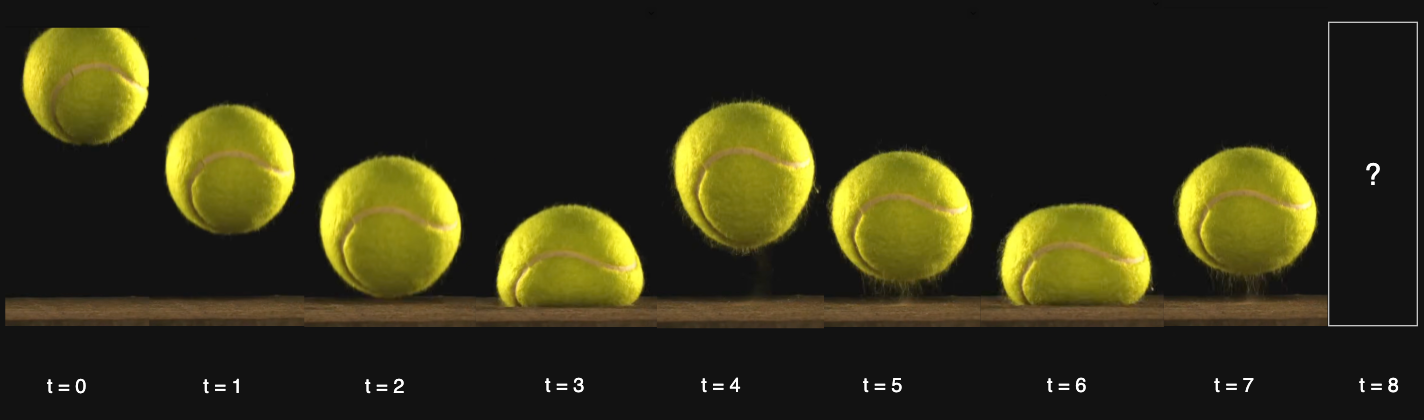}\\[2pt]
{\small (a) Causal time-stepping}\\[6pt]
\includegraphics[width=\linewidth]{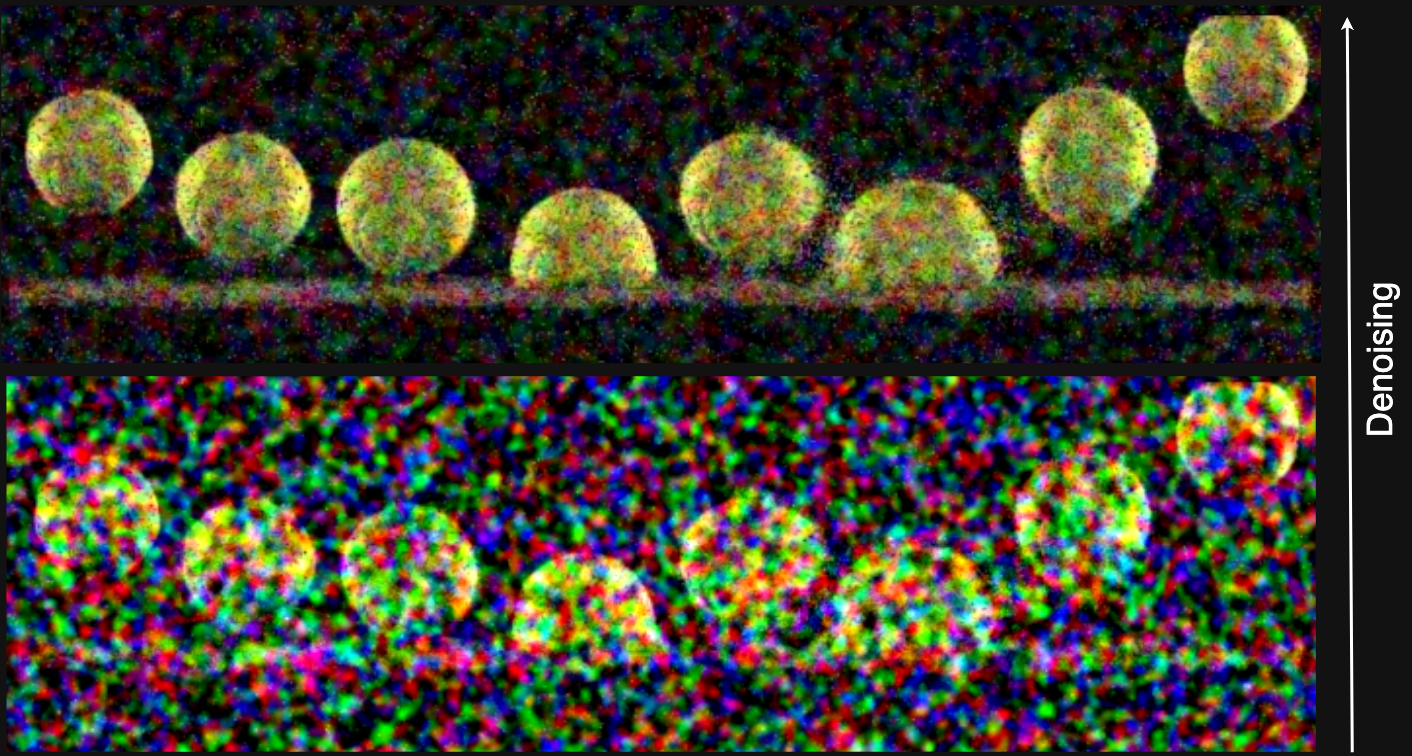}\\[2pt]
{\small (b) Joint collocation}
\caption{\textbf{Two factorizations of the same dynamics.}
\textbf{(a)} Causal generation predicts each chunk from committed history.
\textbf{(b)} Joint generation denoises the whole span at once while the context
is itself still noisy; two states along the trajectory, later toward the top.
(a) is an illustrative rendering; \Cref{sec:evaluation} reports measured physical plausibility on generated video.}
\label{fig:causal-vs-joint}
\end{wrapfigure}

\paragraph{Causal Factorization as a Physics Prior.}

Chunk-wise autoregression is usually motivated by latency and streaming format of the video. In this work, we emphasize that it also matches the structure of the dynamics being generated. Physical evolution is causal and effectively Markovian: contact, momentum transfer, and deformation are determined by the immediately preceding state, so the kernel $p_\theta(z_k\mid H_k,c_k,r_k)$ poses generation as an initial-value problem advanced by a learned transition operator, in which each chunk is a comparatively low-dimensional prediction from history that has already been realized at full precision. This is analogous to the time-stepping approach widely adopted in numerical simulation. On the other hand, segment-level bidirectional generation parameterizes an entire trajectory and denoises it jointly, which is closer to the numerical method space-time collocation: every frame constrains every other frame, and the discrete timing of a contact must be resolved simultaneously with the motion surrounding it (\cref{fig:causal-vs-joint}). The denoising schedule makes this method even more challenging to solve: for most of the joint denoising trajectory the context is itself noisy, so the fine-grained historical trajectory that decides a bounce is unavailable at the point where the surrounding layout is committed. With autoregressive time-stepping, causality is structural rather than statistical---the factorization cannot reference a future chunk, whereas bidirectional temporal attention is symmetric up to positional encoding and must acquire the arrow of time from data during training.

Space–time collocation is a legitimate discretization strategy in numerical methods, and in practice it is adopted deliberately when paired with a robust nonlinear solver. Coupling all space–time dimensions at once, however, yields a nonlinear system that is larger, more strongly entangled, and more sensitive to initialization than the corresponding sequence of single-step solves, and that cost is paid for high temporal order, stiffness tolerance, or parallelism across time steps~\cite{ocfe1975,biegler2010,gander2015}. Bi-directional video denoising inherits the cost structure and buys the last of these---every frame is updated at once on the accelerator---but not the mechanism that makes collocation accurate. A collocation solver earns its cost by iterating until the residual vanishes at chosen points; a few-step amortized denoiser, however, applies a fixed number of approximate updates with no residual to evaluate and no convergence criterion, having been trained to match a score in expectation rather than to enforce a constraint on any particular sample. Within its objective, precisely timed, high-frequency events also carry little weight relative to low-frequency appearance and global layout. Under the proposed chunk-wise autoregression, denoising steps are applied within each short chunk, so this argument bounds the horizon over which a joint solve is attempted rather than eliminating the joint solve altogether. The bidirectional stage in \cref{fig:training-roadmap} remains the source of the appearance and short-range motion prior, which autoregression adaptation re-poses under a causal contract rather than discards. To further enhance the visual plausibility of causal physics, \Cref{sec:post_training} adds a world-model reward that scores each chunk as a successor to its history, and \Cref{sec:evaluation} reports the corresponding physical-plausibility measurements.

\subsection{Generation Model}

\paragraph{Pretraining and Streaming Adaptation.}

We first train on 1--5-second videos with full spatial-temporal attention, then initialize a chunk-wise streaming model from the shared bidirectional weights and adapt it on ordered chunks from 3--10-second videos~\cite{causvid2025,selfforcing2025,longlive2025,causalforcing2026}. Both regimes use the linear rectified-flow objective~\cite{flowmatching2023,rectifiedflow2023}:
\begin{equation}
\begin{aligned}
  \widetilde z_{\sigma}&=(1-\sigma)z+\sigma\epsilon,\qquad
  \epsilon\sim\mathcal{N}(0,I),\\
  \mathcal{L}_{\mathrm{RF}}
  &=\mathbb{E}\!\left[
    \left\|v_{\theta}(\widetilde z_{\sigma},\sigma;q)
    -(\epsilon-z)\right\|_2^2
  \right].
\end{aligned}
\label{eq:unified-flow}
\end{equation}
Here, $\sigma=0$ denotes clean data, $\sigma=1$ pure noise, and generation integrates the learned field from noise toward data. For bidirectional pretraining, $z$ is the complete short-clip latent and $q=c$; for streaming adaptation, $z=z_k$ and $q=(c_k,r_k,H_k)$. Regression errors are averaged over latent elements, streaming losses are first averaged over the chunks in each example, and the outer expectation covers data, noise levels, Gaussian noise, and any sampled history augmentation. Only the current chunk is noised during streaming training; $H_k$ contains committed past content. Shared parameters are initialized from the bidirectional checkpoint, while streaming-specific history and memory parameters are initialized separately.

\paragraph{Bounded Multi-Scale Memory.}
The initial history $H_1$ is supplied by the generation mode. After a chunk is committed, an incremental causal constructor updates the retained history and its fixed-shape persistent state $M_k$ without recompressing the complete prefix. Recent chunks remain at native latent granularity while older spans are progressively compressed under a fixed token budget, following recency-structured and learned-memory designs~\cite{framepack2025,far2025,helios2026,fademem2026,videossm2025,echoinfinity2026}. Orbis reads persistent memory as an additional context tier and writes to it only when detailed history is evicted, preventing denoising evaluations from mutating memory implicitly. Multi-chunk training teaches the state to preserve long-range subject and scene cues while suppressing recurrent drift, with active memory and per-chunk cost independent of rollout length~\cite{echomemory2026}.

\paragraph{Mid-Training.}

To strengthen the model's ability to generate high-quality video, we apply stricter filtering and selection criteria to construct a higher-quality data subset for mid-training. The selected examples emphasize visual fidelity, clear and natural motion, accurate captions, and temporally complete events. The subset is balanced across natural and scientific scenes, people and cultural activity, urban and indoor environments, transport and industrial processes, and creative or digital material. Source groups include professionally captured footage, documentary and educational material, long-form everyday recordings, and diverse mobile or handheld capture conditions. These groups are rebalanced jointly with duration, motion, camera behavior, event density, and language so that the emphasis on quality does not narrow the model to a limited range of visual styles or content.

In the first stage, we train on short videos that each contain a single complete event and are paired with a corresponding caption. This \emph{single-event training} strengthens local image quality, motion naturalness, composition, and prompt following while preserving the history-conditioned objective in \cref{eq:unified-flow}. The second stage uses the event boundaries and interval-local captions described in \cref{sec:data}. During \emph{event-aligned temporal training}, each event condition is activated only over its annotated interval, and successive chunks may receive different instructions within the same continuous video. The model must follow the current event while preserving attributes that should remain stable across the transition. Because these examples extend to 240 seconds, this stage also exposes the model to substantially longer histories, repeated condition changes, and recovery after imperfect intermediate chunks.

\paragraph{Long-Horizon Reliability.}

Long event-aligned examples establish the temporal curriculum, but duration alone does not expose the model to the imperfect histories encountered during autoregressive generation~\cite{diffusionforcing2024,selfforcing2025,framepack2025}. We therefore mix clean histories with two augmentation branches during streaming training. The first corrupts the committed training prefix before both retained history and persistent memory are constructed, using temporally varying degradation, temporal corruption, statistic shifts, and noise~\cite{helios2026,framepack2025}. The second measures fixed-dimensional drift statistics between generated and reference histories matched by initialization, control schedule, and rollout horizon, then reapplies sampled shifts to clean bounded histories~\cite{selfforcing2025,rollingforcing2025}. Both branches are used only after the first chunk, leave $H_1$, the current clean target, and external conditions unchanged, and are included in the expectation of \cref{eq:unified-flow}.

\paragraph{Fine-Tuning.}

The final quality fine-tuning stage starts from clips that pass the strictest data-quality gates. Human reviewers score visual fidelity, motion naturalness, temporal coherence, composition, semantic clarity, subject and background stability, caption correctness, and the clarity of event transitions. Only sequences with an unambiguous, temporally localizable change---such as a person entering the scene, an animal or object appearing, or the camera turning toward a distinct region such as the sky---are retained for event-aligned fine-tuning. Repeated annotation and adjudication are used to construct a compact, high-confidence training subset rather than relying on a single aggregate quality score.

\subsection{Post-Training}
\label{sec:post_training}

\paragraph{Distillation and Reinforcement Learning.}

Post-training first reduces inference cost and then aligns the streaming policy with video-quality rewards. We distill a fixed classifier-free-guidance scale into a conditional-only student by matching a frozen teacher's detached guided prediction at the same noised current-chunk state, causal history, and optional visual condition~\cite{guideddistillation2023}. The scale is embedded in the student weights, so inference requires one conditional evaluation without applying classifier-free guidance again.

Few-step distillation begins only after the chunk-wise causal architecture has been learned. We first apply endpoint-anchored consistency distillation to states on the same frozen reference probability-flow trajectory, using fixed causal context and an exponential-moving-average target branch~\cite{consistencymodels2023,rcm2026,causalrcm2026}. We then refine the model with self-forcing DMD on student-generated histories~\cite{selfforcing2025,causvid2025,causalrcm2026}. At each noised student sample, DMD compares the frozen reference score with an online fake-score model fitted to the student distribution; their difference defines the generator update~\cite{dmd2024}. This ordering preserves causal structure before self-generated distribution matching and avoids directly distilling a bidirectional flow map into a causal student~\cite{causalforcing2026}.

Finally, a frozen old-policy snapshot collects groups of stochastic rollouts for GRPO-style reinforcement learning. Current and old transition likelihoods are evaluated on the same recorded states and realized lower-noise actions under a fixed stochastic-sampler schedule, using the standard clipped objective~\cite{deepseekmath2024,ppo2017}. Visual-quality, motion-quality, and text--video-alignment rewards are normalized separately within each prompt group and combined into a detached relative advantage. Training uses the marginal-preserving stochastic sampler of Flow-GRPO~\cite{flowgrpo2025} and the KL-free multi-reward video variant of DanceGRPO~\cite{dancegrpo2025}; deployment returns to deterministic flow sampling.

\paragraph{Physics-Aware Reward Alignment.}

 \Cref{ssec:formulation} argued that the causal factorization gives generation the
  structure of an initial-value problem; nothing in the training objective,
  however, scores whether a chunk is a plausible successor to the state
  that preceded it. Visual-quality and text-alignment rewards score only
  how a chunk looks. We therefore add
a predictive-consistency reward derived from a latent world model, following the
principle of WMReward~\cite{yuan2026wmreward}. Let $E$ denote a frozen V-JEPA~2
encoder and $P$ its frozen predictor~\cite{vjepa2_2025}. Given the committed
context preceding chunk $k$, the predictor forecasts the representation of the
following interval, and the candidate chunk is embedded by the same encoder:
\begin{equation}
  r_{\mathrm{phys}}(z_k \mid H_k)
  = \operatorname{sim}\!\big(
      P(E(H_k)),\; E(z_k)
    \big),
\label{eq:wm-reward}
\end{equation}
where $\operatorname{sim}(\cdot,\cdot)$ is a normalized agreement in representation
space. Because prediction happens in learned representations rather than pixels,
the signal responds to state transitions---displacement, contact,
deformation---rather than to appearance, and is complementary to the visual- and
motion-quality rewards alongside which it is normalized.

This reward is optimized jointly with distribution matching rather than as a
separate stage, following reward-tilted distribution matching~\cite{rtdmd2026}.
Tilting the teacher distribution $p_\psi$ by a scalar reward $r$ and minimizing the
reverse KL to it separates into two familiar terms,
\begin{equation}
  \nabla_\theta D_{\mathrm{KL}}\!\left(p_\theta \,\|\, \tilde p_\psi\right)
  = \underbrace{\nabla_\theta D_{\mathrm{KL}}\!\left(p_\theta \,\|\, p_\psi\right)}
      _{\text{distribution matching}}
  \;-\; \beta\,\underbrace{\nabla_\theta \mathbb{E}_{p_\theta}\!\left[r\right]}
      _{\text{reward maximization}},
  \qquad
  \tilde p_\psi \propto p_\psi \exp(\beta r),
\label{eq:reward-tilted}
\end{equation}
so few-step distillation and reward optimization descend on one objective and the
teacher's generative prior is retained while the student is steered. Because the
few-step sampler is stochastic at intermediate steps and deterministic at the last,
the reward gradient in \cref{eq:reward-tilted} is estimated in two parts---a
group-relative policy gradient over the stochastic transitions and a pathwise
gradient through the final deterministic step---and noise is shared within a group at
unselected steps so that reward differences are attributed to the steps actually
being updated. The world model is frozen throughout and contributes a scalar to
$r$; it does not backpropagate through its own representations.

\subsection{Super-Resolution}

\paragraph{Video Super-Resolution.}
\label{sec:superres}

The native generator produces semantic and motion structure at $832\times480$. A dedicated single-refinement video super-resolution model transforms each progressively decoded window into 4K output while preserving the motion established by the native generator.

\paragraph{Temporal-Consistency Distillation for Super-Resolution.}

The high-capacity video autoencoder used by the restoration path is too expensive for the end-to-end real-time budget. We therefore develop a two-stage temporally aware distillation procedure to replace it with a compact student, related to recent work on compressed autoencoders and distilled video-VAE decoders~\cite{dcae2024,turbovaed2025,flashvaed2026}. Stage A jointly trains the compact encoder and decoder through a native round trip, in which both compact components process the input video, and a cross-latent path, in which the compact decoder receives latents from the high-capacity encoder. Latent regression and cosine alignment bring the compact representation into the teacher latent space, while pixel and DISTS perceptual reconstruction losses preserve spatial detail~\cite{dists2020}.

Stage B adapts the compact path to the restored latent distribution encountered during super-resolution inference. It first adapts only the compact decoder on restored latents, then fixes that decoder while refining the restoration transformer. The frozen high-capacity teacher and compact student decode the same latent sequence, but the teacher output is used only as a detached target. During transformer refinement, the fixed student decoder remains differentiable with respect to its latent input, so supervision can still reach the restoration transformer; all teacher-derived motion and reliability signals remain detached.

Spatial supervision combines pixel and DISTS perceptual losses against the high-resolution target with reconstruction toward the detached teacher decode~\cite{dists2020}. Temporal supervision matches adjacent student-frame changes to the teacher and adds occlusion-aware motion-compensated consistency~\cite{frvsr2018,tecogan2019}. For each adjacent pair, RAFT is evaluated on the reversed detached teacher-frame pair to obtain the target-to-source field used to backward-warp the earlier student frame onto the later frame's grid~\cite{raft2020}. The corresponding reliability mask is defined on that target grid and broadcast over channels; the residual is averaged over valid pixel-channel entries, with all-invalid transitions omitted. All spatial and temporal terms are averaged over their applicable frames and elements before weighted aggregation.

The restoration transformer uses spatial-window attention while retaining all temporal frames inside each window, following the broader use of temporal propagation and transformer attention in video restoration~\cite{vrt2022,rvrt2022,basicvsrpp2022}. These losses are evaluated within a temporal training chunk and do not introduce an autoregressive dependence between super-resolution chunks. This within-chunk design preserves local restoration capacity and full temporal context while making the model compatible with tiled and distributed execution. Super-resolution improves reconstruction and presentation detail; it is not a mechanism for inventing semantic state absent from the native video. Evaluation reports native-generation and delivered-4K evidence separately.

\section{Inference}
\label{sec:inference}

Visko Orbis~1.0 is served as a continuously active Live Model session. Its inference stack is organized around state reuse, single-stream distributed execution, and progressive video delivery, so that long-form generation and live prompt updates share one steady-state path.

\paragraph{Live State and Conditioning.}
Each session retains a bounded visual history together with its per-layer History KV Cache, history-context features, rotary-position tensors~\cite{roformer2021}, and the active text condition. A rolling prompt summary carries established entities, relationships, environment, and style across interactions. New prompts are encoded asynchronously, admitted at the next uncommitted chunk, and guarded by session and prompt versions so that stale work cannot overwrite the current condition. Generic prefix-aware runtimes reuse immutable attention state, while causal video systems reuse history across autoregressive chunks~\cite{promptcache2023,sglang2024,causvid2025,longlive2025,longlive22026}. Orbis applies these ideas at separate state lifetimes for text cross-attention K/V, prompt embeddings, and compatible visual history.

\paragraph{Compiled Transformer Execution.}
The distilled generator requires one conditional transformer evaluation per denoising state. Orbis compiles the fixed-shape transformer as a whole and combines it with fused QKV projection, AdaLN, Q/K normalization, gated residual, and rotary-position kernels; spatial grids, timestep projections, and communication workspaces are precomputed and reused~\cite{tvm2018,triton2019,pytorch22024,flashattention22024}. The solver further replaces small generic linear-algebra calls with a closed-form UniPC corrector and keeps timestep state on device~\cite{unipc2023}. Shape-specific attention selection, calibrated W8A8 execution, BF16 decoding, reusable buffers, and guarded in-place updates reduce memory traffic while retaining higher precision at sensitive boundaries~\cite{smoothquant2023,viditq2024}.

\paragraph{Full-Sequence Multi-GPU Execution.}
To reduce the latency of a single active stream, video tokens remain sequence-sharded throughout the transformer stack. Self-attention uses Ulysses-style sequence-to-head redistribution, while residual, feed-forward, cross-attention, and conditioning operations remain local to each shard~\cite{deepspeedulysses2023,usp2024,xdit2024,longlive22026}. Q, K, and V are packed into a single all-to-all exchange; fused pack/unpack kernels write directly into reusable communication buffers, and timestep conditioning is materialized only for the local sequence. This full-sequence contract avoids repeatedly reconstructing the global token tensor between blocks and makes parallelism effective for batch-one streaming inference.

\paragraph{Progressive Decode and Delivery.}
The pixel path uses a tensor-only compiled video decoder with temporal assembly buffers, a convolution-friendly three-dimensional memory layout, and tiled or spatially parallel execution when required. Latent generation, decoding, color conversion, media encoding, and delivery are connected through bounded FIFO queues and device events, allowing each completed temporal group to move downstream without retaining the full video on the GPU. Model weights, compiled graphs, and workspaces remain resident across chunks and requests, while asynchronous copies and background media assembly keep host synchronization outside the generation path. Similar decoder and pipeline co-design has become central to recent real-time video systems and block-wise video autoencoders~\cite{seedance2025,vidus12026,longlive22026,wfvae2024}.

\paragraph{Content-Adaptive Drift Stabilization.}
Long rollouts can expose errors that are weak in any individual chunk but accumulate through recurrent history. Orbis therefore provides a conservative set of inference-time controls: training-aligned additive perturbation of selected pre-compression history latents, temporal RoPE scaling, history-attention reweighting, periodic VAE round-trip refresh, segment-local latent-statistics correction, and motion-aware negative-prompt routing. These controls target complementary forms of over-trusted history, motion escalation, and color drift, drawing on the broader literature on noisy causal context, bounded long-video memory, and temporal-frequency adaptation~\cite{diffusionforcing2024,longlive2025,rollingforcing2025,framepack2025,memflow2025,fademem2026,riflex2025}. Because no fixed combination is safe across people, object, low-motion, and high-motion content, they are organized as content-specific profiles rather than a universal default; the unmodified path remains the correctness and quality reference.

\paragraph{Physics-Aware Inference Alignment.}
We additionally use a latent-world-model reward to improve physical plausibility at inference time. Following the principle of WMReward~\cite{yuan2026wmreward}, the world model predicts future representations from the visual context and compares them with candidate frames; the resulting predictive-consistency score steers or selects denoising trajectories toward plausible motion and object dynamics.

\paragraph{Streaming Super-Resolution.}
Native $832\times480$ chunks are converted to 4K by a single-refinement video super-resolution model. A temporally distilled compact autoencoder limits decode cost, while spatial-window attention preserves the complete temporal extent within each restoration window~\cite{vrt2022,rvrt2022,basicvsrpp2022}. Window-shard sequence parallelism, halo-aware tiled decoding, and online merging of distributed softmax statistics provide the corresponding execution path~\cite{onlinesoftmax2018,flashattention22024}. Together with progressive assembly, the generator and super-resolution stages deliver 4K video at 24 FPS without waiting for the complete sequence.

\section{Evaluation}
\label{sec:evaluation}

\paragraph{Evaluation protocol.}

We evaluate real-time interactive long-video generation on 74 cases~\cite{narrlv2025,directorbench2026}. Each case contains six events. Depending on the content richness of the case, each event spans 10, 20, or 30 seconds, yielding videos of approximately one, two, or three minutes, respectively. We score every event against the prompt active during that interval and average each metric across all events in the benchmark.

\paragraph{Baselines and metrics.}

We compare with representative real-time, autoregressive, and long-video systems: Helios~\cite{helios2026}, Self-Forcing~\cite{selfforcing2025}, Rolling-Forcing~\cite{rollingforcing2025}, Causal-Forcing~\cite{causalforcing2026}, LongLive~\cite{longlive2025}, MemFlow~\cite{memflow2025}, InfinityStar~\cite{infinitystar2025}, and SANA-Video~\cite{sanavideo2025}. Following multidimensional video-evaluation practice, the protocol separates perceptual quality, motion quality, prompt alignment, and distributional diagnostics rather than relying on a single aggregate score~\cite{vbenchpp2024,evalcrafter2024,fetv2023}.

DOVER (Disentangled Objective Video Quality Evaluator)~\cite{dover2023} provides no-reference aesthetic and technical quality scores. For VideoAlign/VideoReward~\cite{videoalign2025}, we sample events at two frames per second and obtain unbounded logits for visual quality, motion quality, and text alignment; their sum is the overall reward. For Human Preference Score v3 (HPSv3)~\cite{hpsv32025}, we score seven uniformly sampled frames and report mean and minimum frame rewards. HeliosBench reports semantic alignment from a video-language contrastive encoder, aesthetics from an image-language predictor trained on LAION aesthetic ratings, and net directional motion from signed mean Farneb\"ack optical flow~\cite{helios2026,farneback2003}.

\paragraph{Quantitative results.}

On the prompt-independent dimensions in Table~\ref{tab:longvideo-quality}, Orbis leads DOVER aesthetic and technical quality (0.8101 and 0.5572) and VideoAlign visual and motion quality (1.5777 and 1.8646). Its HeliosBench aesthetic score, 0.5954, ranks third behind SANA-Video and LongLive.

On the prompt-aware dimensions, Orbis has the largest raw VideoAlign text-alignment logit (0.1043), VideoAlign overall reward (3.5466), and HeliosBench semantic-alignment score (0.2361). Its Human Preference Score v3 frame mean and minimum increase to 8.1018 and 6.9719, respectively, placing both raw values behind SANA-Video and LongLive but ahead of MemFlow.

\begin{table}[H]
\centering
\caption{Quality, alignment, and motion measurements on the one- to three-minute event-based benchmark. Dark, medium, and pale green mark the first, second, and third raw values in each quality column. Panel B contains prompt-aware metrics; its coloring is descriptive rather than a formal cross-system rank because prompt suites differ. VideoAlign and HPSv3 outputs are unbounded reward logits. Motion is descriptive and has no preferred direction.}
\label{tab:longvideo-quality}
\scriptsize
\setlength{\tabcolsep}{8.2pt}
\textit{Panel A: Prompt-independent quality and motion}\par\vspace{2pt}
\begin{tabular*}{\linewidth}{@{\extracolsep{\fill}}lcccccc}
\toprule
System & \makecell{DOVER\\Aesthetic $\uparrow$} & \makecell{DOVER\\Technical $\uparrow$} & \makecell{VideoAlign\\Visual Quality $\uparrow$} & \makecell{VideoAlign\\Motion Quality $\uparrow$} & \makecell{HeliosBench\\Aesthetic $\uparrow$} & \makecell{HeliosBench\\Motion Amplitude} \\
\midrule
Helios & 0.4989 & 0.2123 & \ranksecond{1.3881} & 1.5714 & 0.5329 & 0.1782 \\
Self-Forcing & 0.7057 & 0.3983 & \rankthird{1.3739} & \ranksecond{1.8531} & 0.5819 & 0.0456 \\
SANA-Video & 0.7913 & 0.4752 & 1.3220 & \rankthird{1.7239} & \rankfirst{0.6153} & 0.1260 \\
Rolling-Forcing & 0.7702 & 0.4669 & 1.2826 & 1.5650 & 0.5596 & 0.0894 \\
LongLive & \ranksecond{0.8095} & \ranksecond{0.5166} & 1.2297 & 1.6667 & \ranksecond{0.6016} & 0.0973 \\
MemFlow & \rankthird{0.7969} & \rankthird{0.4986} & 1.2130 & 1.6231 & 0.5842 & 0.1451 \\
InfinityStar & 0.7589 & 0.4884 & 1.1905 & 1.6102 & 0.5190 & 0.4516 \\
Causal-Forcing & 0.7527 & 0.4663 & 1.0799 & 1.4060 & 0.5321 & 1.2283 \\
\midrule
\model & \rankfirst{0.8101} & \rankfirst{0.5572} & \rankfirst{1.5777} & \rankfirst{1.8646} & \rankthird{0.5954} & 0.2477 \\
\bottomrule
\end{tabular*}
\par\vspace{12pt}
\textit{Panel B: Prompt-aware alignment and frame preference}\par\nointerlineskip\vspace{2pt}
\begin{tabular*}{\linewidth}{@{\extracolsep{\fill}}lccccc}
\toprule
System & \makecell{VideoAlign\\Text Alignment $\uparrow$} & \makecell{VideoAlign\\Overall Reward $\uparrow$} & \makecell{HPSv3\\Frame Mean $\uparrow$} & \makecell{HPSv3\\Frame Minimum $\uparrow$} & \makecell{HeliosBench\\Semantic Alignment $\uparrow$} \\
\midrule
Helios & \rankthird{-0.1431} & 2.8164 & 3.2673 & 1.6807 & 0.1620 \\
Self-Forcing & \ranksecond{-0.1406} & \ranksecond{3.0864} & 5.6802 & 4.5280 & 0.1727 \\
SANA-Video & -0.2201 & \rankthird{2.8258} & \rankfirst{8.4373} & \rankfirst{7.5114} & 0.1963 \\
Rolling-Forcing & -0.2345 & 2.6131 & 7.4059 & 6.3367 & 0.1882 \\
LongLive & -0.2366 & 2.6597 & \ranksecond{8.3717} & \ranksecond{7.2633} & \ranksecond{0.2036} \\
MemFlow & -0.2445 & 2.5915 & 8.0721 & 6.9502 & \rankthird{0.2030} \\
InfinityStar & -0.2263 & 2.5744 & 4.4570 & 2.9372 & 0.1842 \\
Causal-Forcing & -0.3137 & 2.1722 & 6.3071 & 3.5434 & 0.2017 \\
\midrule
\model & \rankfirst{0.1043} & \rankfirst{3.5466} & \rankthird{8.1018} & \rankthird{6.9719} & \rankfirst{0.2361} \\
\bottomrule
\end{tabular*}
\end{table}

\paragraph{Physical plausibility evaluation.}

 \Cref{ssec:formulation} argued that causal chunk-wise generation matches the
  structure of physical dynamics, and \Cref{sec:post_training} added a world-model
  reward to supervise it. Visual quality and prompt alignment do not by
  themselves establish whether that holds at the delivered-video boundary. We therefore use three complementary protocols: a long-video assessment with two video-based physical-reasoning evaluators, the controlled Physics-IQ image-to-video benchmark, and the fine-grained Physics track of VBench-2.0~\cite{videophy22025,cosmosreason12025,physicsiq2026,vbench22025}.

For the long-video assessment, we use the 60-second scenes from the Arena suite and divide every output into non-overlapping 10-second segments. VP2-PC is the caption-free, video-only physical-commonsense head of VideoPhy-2 AutoEval and returns a score from 1 to 5~\cite{videophy22025}. We report both its mean score and the fraction of segments scoring at least 4. As a second evaluator, we use Cosmos-Reason1-7B with the official Cosmos Cookbook physics rubric, sampling each segment at 16 frames per second and again scoring from 1 to 5~\cite{cosmosreason12025,cosmoscookbook2026}. Table~\ref{tab:longvideo-physics} reports the system-level results.

\begin{table}[H]
\centering
\caption{Physical plausibility on long-video Arena scenes. Each 60-second output is divided into non-overlapping 10-second segments. VP2-PC and Cosmos scores range from 1 to 5; higher is better. Dark, medium, and pale green mark the first, second, and third scores in each metric column.}
\label{tab:longvideo-physics}
\scriptsize
\setlength{\tabcolsep}{9.0pt}
\begin{tabular*}{\linewidth}{@{\extracolsep{\fill}}lccc}
\toprule
System & \makecell{VP2-PC\\Mean $\uparrow$} & \makecell{VP2-PC\\$\mathrm{PC}\!\geq\!4$ (\%) $\uparrow$} & \makecell{Cosmos\\Mean $\uparrow$} \\
\midrule
\model & \rankfirst{4.284} & \rankfirst{89.8} & \rankfirst{3.685} \\
Causal-Forcing & \ranksecond{4.280} & 85.2 & 3.596 \\
Helios & \rankthird{4.277} & 83.7 & \ranksecond{3.670} \\
LongLive-2 & 4.254 & 86.7 & \rankthird{3.668} \\
HappyOyster & 4.235 & \rankthird{88.3} & 3.481 \\
Rolling-Forcing & 4.167 & 80.3 & 3.519 \\
Krea Realtime & 4.163 & 82.6 & 3.096 \\
Odyssey-2 & 4.133 & \ranksecond{89.4} & 2.927 \\
PixVerse-R1 & 3.822 & 68.6 & 3.031 \\
\bottomrule
\end{tabular*}
\end{table}

\model achieves the highest mean score under both evaluators, with 4.284 on VP2-PC and 3.685 on Cosmos, and also the largest fraction of high-scoring VP2-PC segments at 89.8\%. The margins in the mean scores are narrow---Causal-Forcing reaches 4.280 on VP2-PC, while Helios and LongLive-2 reach 3.670 and 3.668 on Cosmos---but the two independently configured evaluators agree on the top system.

For Physics-IQ~\cite{physicsiq2026}, we follow the official single-frame-conditioned image-to-video protocol. Generated continuations are encoded and trimmed to exactly five seconds before evaluation with the original benchmark ground truth. We evaluate the 24-FPS variants of \model and Helios and the 16-FPS variants of the streaming baselines, reconstructing ground-truth temporal alignment with the benchmark's official linear-interpolation rule. The aggregate Physics-IQ score combines spatial IoU, weighted spatial IoU, spatiotemporal IoU, and mean squared error (MSE); the component scores are included to make the trade-offs visible.

\begin{table}[H]
\centering
\caption{Physics-IQ results for single-frame-conditioned image-to-video continuations. Higher is better except for MSE. Dark, medium, and pale green mark the first, second, and third scores in each column. $^\dagger$Self-Forcing has no official image-to-video weights; we approximate this mode using its EMA generator, independent first-frame conditioning, and an expanded KV cache.}
\label{tab:physics-iq}
\scriptsize
\setlength{\tabcolsep}{7.0pt}
\begin{tabular*}{\linewidth}{@{\extracolsep{\fill}}lccccc}
\toprule
System & \makecell{Physics-IQ\\$\uparrow$} & \makecell{Spatial\\IoU $\uparrow$} & \makecell{Weighted\\Spatial $\uparrow$} & \makecell{Spatiotemporal\\IoU $\uparrow$} & \makecell{MSE\\$\downarrow$} \\
\midrule
\model & \rankfirst{26.93} & \ranksecond{0.296} & \ranksecond{0.211} & \rankfirst{0.316} & \rankfirst{0.0050} \\
Helios & \ranksecond{26.78} & \rankthird{0.282} & \rankfirst{0.255} & \ranksecond{0.193} & \ranksecond{0.0089} \\
InfinityStar & \rankthird{20.36} & \rankfirst{0.330} & \rankthird{0.194} & 0.115 & \rankthird{0.0091} \\
Self-Forcing$^\dagger$ & 15.46 & 0.191 & 0.144 & \rankthird{0.175} & 0.0154 \\
Causal-Forcing & 9.10 & 0.202 & 0.103 & 0.072 & 0.0347 \\
\bottomrule
\end{tabular*}
\end{table}

On Physics-IQ, \model obtains the highest aggregate score at 26.93, narrowly exceeding Helios at 26.78. It also leads spatiotemporal IoU at 0.316 and attains the lowest MSE at 0.0050. InfinityStar records the highest spatial IoU, while Helios leads weighted spatial IoU, showing why the aggregate and component metrics should be read together. These controlled continuation results complement the long-video evaluator scores: the former compares generated motion with recorded physical outcomes, whereas the latter measures perceived physical commonsense throughout extended outputs.

For VBench-2.0 Physics~\cite{vbench22025}, we evaluate the Mechanics, Thermotics, and Material prompt groups. We follow the official \texttt{vbench\_standard} evaluation pipeline: LLaVA-Video-7B-Qwen2 judges uniformly sampled frames using the benchmark-provided question--answer pairs, first performing the benchmark validity check and then assigning a binary physics-correctness score to each valid video. Dimension scores average the valid binary responses, and the overall mean is the unweighted average across the three dimensions. For \model, we use the four-step distilled configuration through its structured HTML prompting interface; all comparison systems use the official English prompts. Because the structured \texttt{scene} field can make the expected physical outcome explicit, particularly for Material, this is an end-to-end comparison of model-plus-prompting configurations rather than a controlled model-only ranking.

\begin{table}[H]
\centering
\caption{VBench-2.0 Physics results. Higher is better. Invalid videos are excluded following the official protocol. Dark, medium, and pale green mark the first, second, and third scores in each column.}
\label{tab:vbench2-physics}
\scriptsize
\setlength{\tabcolsep}{10.0pt}
\begin{tabular*}{\linewidth}{@{\extracolsep{\fill}}lcccc}
\toprule
System & Mean $\uparrow$ & Mechanics $\uparrow$ & Thermotics $\uparrow$ & Material $\uparrow$ \\
\midrule
\model & \rankfirst{0.7125} & \rankfirst{0.7021} & \rankfirst{0.6966} & \rankfirst{0.7387} \\
InfinityStar & \ranksecond{0.5773} & \ranksecond{0.6736} & \ranksecond{0.5000} & \ranksecond{0.5583} \\
Self-Forcing & \rankthird{0.4925} & 0.5520 & 0.4672 & \rankthird{0.4583} \\
SANA-Video & 0.4717 & 0.5556 & 0.4929 & 0.3667 \\
LongLive & 0.4678 & 0.5667 & 0.4779 & 0.3587 \\
Helios & 0.4624 & \rankthird{0.5812} & 0.4690 & 0.3370 \\
Rolling-Forcing & 0.4600 & 0.5000 & \rankthird{0.4964} & 0.3837 \\
Causal-Forcing & 0.4540 & 0.5426 & 0.4380 & 0.3814 \\
MemFlow & 0.4484 & 0.5333 & 0.4853 & 0.3265 \\
\bottomrule
\end{tabular*}
\end{table}

Under these evaluated system configurations, \model obtains the highest overall mean (0.7125) and the highest score in every physics dimension, including 0.7021 in Mechanics, 0.6966 in Thermotics, and 0.7387 in Material. InfinityStar is the strongest comparison system in the mean and all three dimensions, while Helios records the third-highest Mechanics score. The cross-protocol prompt difference described above should be kept in view when interpreting these gaps.

\paragraph{Human preference evaluation.}

Automated rewards do not fully capture whether a long video remains convincing as events change and generation continues~\cite{fetv2023}. We therefore complement the metric suite with a randomized side-by-side Arena study, following the multidimensional human-evaluation structure used in recent comprehensive video evaluations and the pairwise Arena protocol~\cite{seedance2025,vbenchpp2024,chatbotarena2024}. The study compares eight systems on both 60-second and 120-second outputs. Raters separately select the stronger video in overall preference, visual fidelity, instruction and event-switch compliance, and temporal stability. We aggregate the pairwise decisions into an Elo-scale rating independently for each dimension and report the point estimates.

\begin{table}[H]
\centering
\caption{Multidimensional human preference on long-form video generation. Elo-scale point estimates are computed independently within each dimension; higher is better. Dark, medium, and pale green mark the first, second, and third scores in each column.}
\label{tab:userstudy-arena}
\scriptsize
\setlength{\tabcolsep}{7.0pt}
\begin{tabular*}{\linewidth}{@{\extracolsep{\fill}}lcccc}
\toprule
System & \makecell{Overall\\Elo $\uparrow$} & \makecell{Visual fidelity\\Elo $\uparrow$} & \makecell{Instruction / switch\\Elo $\uparrow$} & \makecell{Temporal stability\\Elo $\uparrow$} \\
\midrule
\textbf{VISKO ORBIS 1.0} & \rankfirst{\textbf{1838}} & \ranksecond{\textbf{1843}} & \ranksecond{\textbf{1711}} & \rankfirst{\textbf{1940}} \\
HappyOyster~\cite{happyoyster2026} & \ranksecond{1734} & \rankfirst{1937} & \rankfirst{1722} & \ranksecond{1807} \\
LongLive-2~\cite{longlive22026} & \rankthird{1562} & 1566 & \rankthird{1525} & 1556 \\
Odyssey-2~\cite{odyssey22025} & 1552 & 1523 & 1507 & \rankthird{1735} \\
PixVerse-R1~\cite{pixverser12026} & 1540 & \rankthird{1637} & 1503 & 1531 \\
Rolling-Forcing~\cite{rollingforcing2025} & 1382 & 1333 & 1376 & 1419 \\
Krea Realtime~\cite{krea2025} & 1323 & 1293 & 1420 & 1169 \\
Causal-Forcing~\cite{causalforcing2026} & 1262 & 1270 & 1377 & 1238 \\
\bottomrule
\end{tabular*}
\end{table}

\paragraph{Comparative analysis.}

The human study reveals a complementary pattern to the automated evaluation. \model obtains the highest point estimates for overall preference and temporal stability, while HappyOyster obtains the highest visual-fidelity score and a narrowly higher instruction score. The separation is largest in temporal stability, where \model scores 1940 compared with 1807 for the next-ranked system. Together, the dimension-level results suggest that Orbis's perceived advantage is concentrated in maintaining a coherent long-form viewing experience, while the strongest competing system remains highly competitive on local appearance and instruction compliance.

\begin{figure}[H]
\centering
\includegraphics[width=0.7\linewidth]{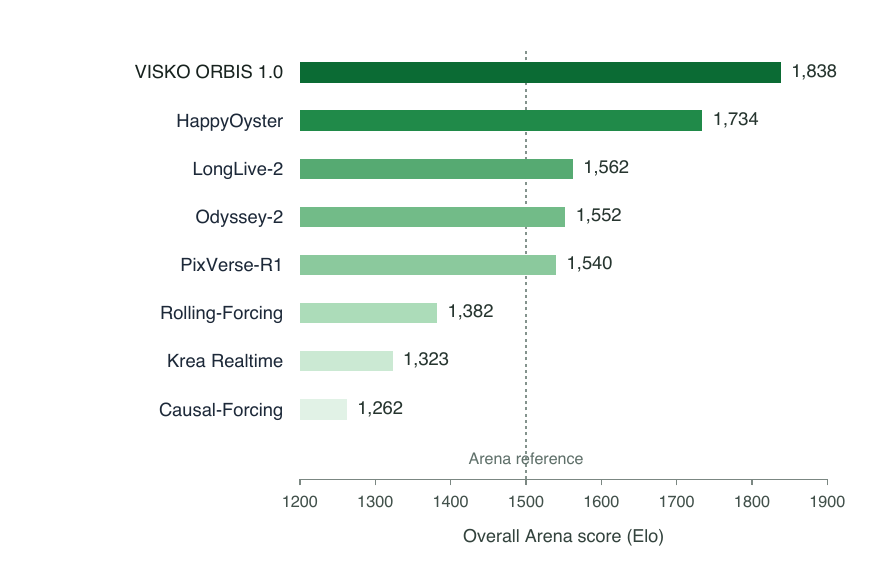}
\caption{Overall long-form video Arena point estimates from the human preference study. \textbf{VISKO ORBIS 1.0} obtains the highest overall score; Table~\ref{tab:userstudy-arena} provides the dimension-level comparison and citations for all baselines.}
\label{fig:userstudy-arena}
\end{figure}

\section{Conclusion}

We presented \model, a Live Model that reframes video generation from producing bounded clips to sustaining a controllable visual stream. By combining stateful chunk-autoregressive generation with event-aligned training, efficient distillation, progressive decoding, and streaming super-resolution, \model supports T2V, I2V, video continuation, and in-generation prompt updates, delivering 4K video at 24 FPS with an average visible response below one second in the reported serving configuration. Automated, human-preference, physical-plausibility, and long-horizon evaluations demonstrate strong visual quality, temporal stability, physically credible dynamics and interactive control. We hope this work helps advance video generation toward a responsive creative medium that users can direct continuously as it unfolds.

\section*{Contributors}

\textbf{Project Leads}: Zhengzhong Tu, Jie Yang, Qing Will Yin

\textbf{Core Contributors}: Xiangbo Gao, Siyuan Yang, Ping He, Mingyang Wu, Yuheng Wu, Yushen Zuo, Jiongze Yu, Ryan Cui, Hongyuan Hua, Devin Ma, Xiao Jin, Yubo Ruan

\textbf{Contact}: xiangbogaobarry@gmail.com, info@visko.ai

{\footnotesize
\setlength{\bibsep}{1pt}
\bibliographystyle{IEEEtran}
\bibliography{references}
}

\end{document}